\documentclass[acmtog]{acmart}
\usepackage{float}
\usepackage{amsmath}

\DeclareMathOperator*{\argmin}{arg\,min}
\usepackage{subcaption}
\AtBeginDocument{%
  }

\copyrightyear{2026}
\acmYear{2026}
\setcopyright{cc}
\setcctype{by}
\acmConference[SA Conference Papers '26]{SIGGRAPH Asia 2026 Conference Papers}{December 01--04, 2026}{Kuala Lumpur, Malaysia}
\acmBooktitle{SIGGRAPH Asia 2026 Conference Papers (SA Conference Papers '26), December 01--04, 2026, Kuala Lumpur, Malaysia}
\acmDOI{10.1145/3829340.3842152}
\acmISBN{979-8-4007-2842-6/2026/12}

\begin{document}

\title{PART: Learning 3D Part Assembly and Retrieval with Transformers}

\author{Ruchao Bao}
\orcid{https://orcid.org/0009-0002-4536-1953}
\affiliation{
  \institution{University of Science and Technology of China}
  \country{China}
}
\email{iambrc@mail.ustc.edu.cn}

\author{Wenzheng Wu}
\orcid{https://orcid.org/0009-0005-4192-8176}
\affiliation{
  \institution{University of Science and Technology of China}
  \country{China}
}
\email{wuwzh@mail.ustc.edu.cn}

\author{Chucheng Xiang}
\orcid{https://orcid.org/0009-0000-3310-5528}
\affiliation{
  \institution{University of Science and Technology of China}
  \country{China}
}
\email{xcc2020@mail.ustc.edu.cn}

\author{Zhongyuan Liu}
\orcid{https://orcid.org/0000-0003-1601-0038}
\affiliation{
  \institution{Tencent}
  \country{China}
}
\email{lockliu@tencent.com}

\author{Yuan Liu}
\orcid{https://orcid.org/0000-0003-2933-5667}
\affiliation{
  \institution{Hong Kong University of Science and Technology}
  \country{China}
}
\email{yuanly@ust.hk}

\author{Jinxin Dong}
\orcid{https://orcid.org/0009-0006-6201-2690}
\affiliation{
  \institution{Tencent}
  \country{United States of America}
}
\email{jxdong@global.tencent.com}

\author{Ligang Liu}
\orcid{https://orcid.org/0000-0003-4352-1431}
\affiliation{
  \institution{University of Science and Technology of China}
  \country{China}
}
\email{lgliu@ustc.edu.cn}

\author{Ziqi Wang}
\orcid{https://orcid.org/0000-0002-3817-3922}
\authornote{Corresponding author.}
\affiliation{
  \institution{Hong Kong University of Science and Technology}
  \country{China}
}
\email{ziqiw@ust.hk}

\begin{abstract}
    3D assembly is fundamental to modern manufacturing and digital content creation. In this paper, we present PART, a unified transformer-based framework for 3D part retrieval and assembly: given a target shape and a part library, PART automatically selects the appropriate parts and predicts their 6-DoF poses to reconstruct the target. While prior work has achieved impressive progress on assembling a pre-defined set of parts, this more practical retrieval-based setting remains largely unexplored. The task faces three key challenges: (i) a combinatorially explosive search space that grows exponentially with library size; (ii) variable-length outputs, as different targets require different numbers of parts; and (iii) continuous 6-DoF pose estimation for part assembly. To address these, we formulate retrieval and assembly as a set prediction problem and design a novel transformer-based framework that retrieves parts and regresses their poses with variable-length output. Additionally, we exploit the duality between part pose estimation and target segmentation through joint training and a novel segmentation-enhanced optimization module. Finally, We curate a large-scale dataset of 80K+ shapes, and the results show that PART generalizes to scene layouts, image targets, and real-world scans. Project Page: \url{https://iambrc.github.io/PART-project-page/}.
\end{abstract}

\ccsdesc[500]{Computing methodologies~Point-based models}
\ccsdesc[300]{Computing methodologies~Shape analysis}
\ccsdesc[100]{Computing methodologies~Neural networks}

\keywords{3D Part Assembly, Point Cloud Understanding, Part-based Modeling, Transformers}
\begin{teaserfigure}
  \includegraphics[width=\textwidth]{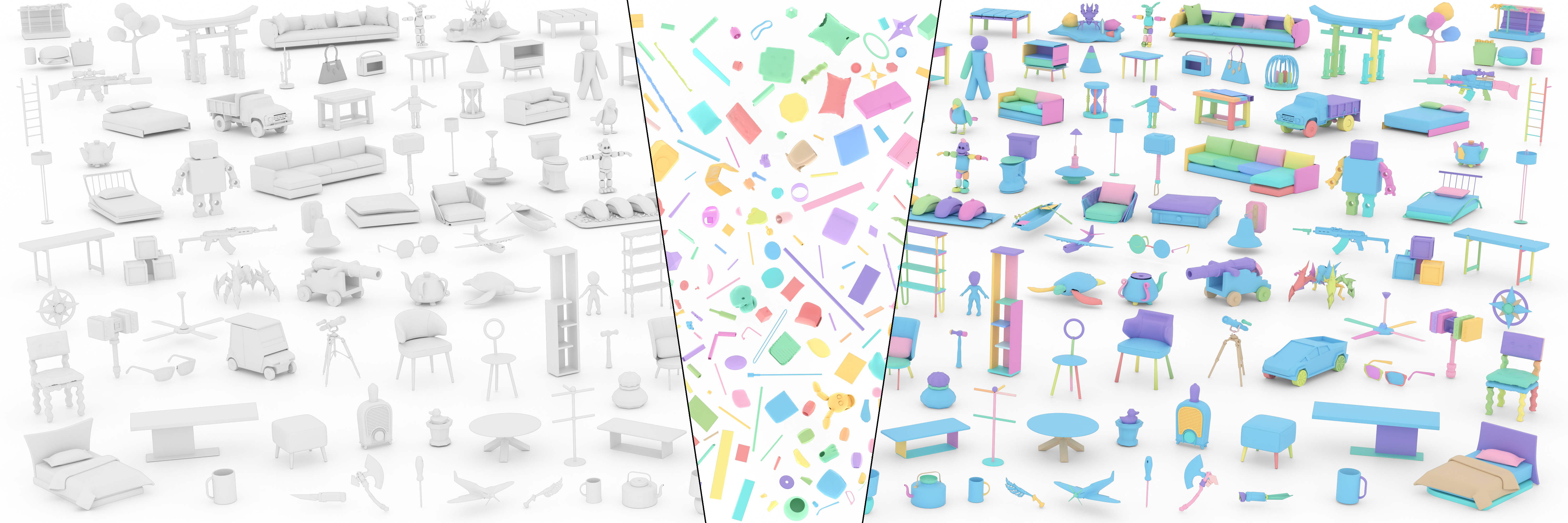}
  \caption{Results of our PART framework. Given a monolithic target shape (left, visualized as a mesh and represented as a point cloud for model input), PART retrieves appropriate components from a candidate part library (middle) and reconstructs the target as a part assembly by predicting the 6-DoF pose of each selected component (right).}
  \Description{Teaser.}
  \label{fig:teaser}
\end{teaserfigure}

\maketitle

\section{Introduction}
Assemblies are fundamental to 3D content creation, gaming, and manufacturing, where complex objects are often represented as compositions of smaller and simpler parts. Traditionally, modeling 3D assemblies has been a labor-intensive process, requiring designers to manually arrange hundreds of components in CAD software. Recent advances in artificial intelligence have shown a promising path toward automating this process, enabling 3D assembly generation from text prompts~\cite{pun2025generating} or reference images~\cite{zhao2025assembler}. Although recent 3D generative models can already synthesize plausible, high-quality geometry from scratch~\cite{lai2025hunyuan3d}, many CAD, industrial design, gaming, and asset-production workflows require more than plausible geometry: objects should be built from known, reusable components so that part identity, editability, and compatibility with existing libraries are preserved. Assembling target shapes from such components library therefore remains an important problem.

Extensive prior work on part assembly has achieved impressive results~\cite{zhan2020generative, xu2025spaformer, zhao2025assembler}. 
However, these methods primarily focus on arranging a predefined set of components, assuming that the provided parts exactly match those required to reconstruct the target shape, which limits their applicability in practical settings where component libraries often contain missing, redundant, or interchangeable parts. 
In both real-world manufacturing and virtual asset creation, the ability to automatically select and reuse parts from a component library is crucial for scalable assembly automation, yet this retrieval-based assembly setting remains largely unexplored. Motivated by this gap, we target a more challenging problem to automate the assembly pipeline: given a target shape and a part library, our system jointly selects appropriate parts and estimates their 6-DoF poses to reconstruct the target geometry.

Part retrieval and assembly face three main challenges. \textbf{(i) Combinatorial explosion:} selecting appropriate parts from a large part library induces a vast search space whose complexity grows rapidly with library size. \textbf{(ii) Variable-length outputs:} different target shapes require different numbers of components, posing a significant challenge for conventional deep networks that typically assume fixed-size outputs. \textbf{(iii) Continuous 6-DoF pose estimation:} unlike modular systems like LEGO~\cite{xu2025legoace}, general part assembly requires reasoning over both semantic and geometric properties of parts to estimate poses in the continuous, non-Euclidean space $\mathrm{SE}(3)$.

We address these challenges through a unified framework built on two key insights.
For the first two challenges, we cast part retrieval and assembly as a \textbf{set prediction} problem (i.e., predicting an unordered collection of items whose exact number can vary). Inspired by DETR~\cite{carion2020end}, we adopt a query-based transformer architecture in which a fixed set of learnable query vectors attends to both the target shape and the candidate parts in a shared feature space. Each query independently decides which part to retrieve and where to place it, while unmatched queries are suppressed, enabling variable-length output within a fixed-capacity model. 
Our approach can handle part libraries containing over $100$ parts within seconds at inference time, since the model processes all queries in parallel in a single forward pass. However, directly training on large libraries is both memory- and time-intensive due to the quadratic complexity of the Transformer architecture. To address this, we adopt a \emph{train-short-test-long} strategy: we first train the model on small libraries augmented with randomly sampled distractor parts, and then apply it at inference time to full-sized libraries (see \autoref{subsec:details}).

For the third challenge, our key observation is that \textbf{assembly and segmentation are dual problems}: the part pose implicitly determines which region of the target it occupies, while a per-part segmentation of the target directly constrains where and how that part is placed. We exploit this duality by jointly training our framework on retrieval, assembly, and segmentation. At test time, we further introduce a segmentation-enhanced optimization module that uses the predicted segmentation masks to refine the initial pose estimates via part-to-region alignment. Moreover, we adopt a continuous rotation representation (A-Matrix) for singularity-free orientation regression and design a symmetry-aware loss that accounts for geometric ambiguities in symmetric parts, jointly mitigating the difficulty of pose estimation in $\mathrm{SE}(3)$. 

Our contributions are summarized as follows: (1) We innovatively formulate 3D part retrieval and assembly as a set prediction problem, and propose PART, a unified transformer-based framework that jointly addresses retrieval and assembly with variable-length output. (2) We introduce a novel segmentation-enhanced optimization module and a symmetry-aware loss to improve assembly accuracy. (3) We curate a large-scale and diverse dataset of 80K+ shapes for general part assembly task, and demonstrate that PART generalizes to scene layouts, image targets, and real-world scans.
\section{Related Work}

\subsection{3D Part Assembly}
The 3D part assembly task aims to assemble multiple parts into a coherent and complete model. Early works employ intelligent scissoring techniques~\cite{funkhouser2004modeling} and probabilistic graphical models~\cite{chaudhuri2011probabilistic, jaiswal2016assembly, kalogerakis2012probabilistic} to address the challenges associated with the semantic and geometric relationships among parts. With the advent of deep learning and the availability of large-scale part-level datasets with detailed annotations~\cite{chang2015shapenet, mo2019partnet}, recent methods have increasingly focused on fully automated assembly through neural networks.
ImagePA~\cite{li2020learning} first proposes single-image-based unlabeled 3D part assembly, and DGL~\cite{zhan2020generative} innovatively proposes an iterative graph neural network to learn parts' relationships and poses. Building on these advances, RGL~\cite{narayan2022rgl} and SPAFormer~\cite{xu2025spaformer} incorporate input part order to improve the assembly performance; IET~\cite{zhang20223d} presents an instance-encoded Transformer framework to better distinguish geometrically similar parts; ScorePA~\cite{cheng2023score} proposes a score-based framework to learn the conditional probability distribution of part poses; 3DHPA~\cite{du2024generative} develops a part-whole-hierarchy message passing network; and Imagine~\cite{wang2025imagine} utilizes a structural knowledge graph to guide the assembly process;
The Shape Part Slot Machine~\cite{wang2022shape} formulates assembly through explicit contact-based reasoning, and CFPA~\cite{zhang2026coarse} introduce a coarse-to-fine framework utilizing semantic super-parts and symmetry-aware pose estimation.
A parallel line of research explores integrating physical constraints (PhysFit~\cite{wang2024physfit} and JointPA~\cite{li2024category}), segmentation-based assembly (GPAT~\cite{li2023rearrangement}), and retrieval-based assembly (UPRA~\cite{xu2023unsupervised}). More recent works focus on advanced models or real-world automated assembly. For example, Assembler~\cite{zhao2025assembler} utilizes an anchor point diffusion model, Rectified Point Flow~\cite{sun2025rectified} and GARF~\cite{li2025garf} use flow matching models to indirectly predict the pose of each part. 
Meanwhile, ManualPA~\cite{zhang2025manual}, Manual2Skill~\cite{tie2025manual}, Manual2Skill++~\cite{tie2025manual2skill++} explore part assembly schemes guided by instruction manuals. 

Our method is most related to UPRA~\cite{xu2023unsupervised} and GPAT~\cite{li2023rearrangement}. UPRA couples a pre-trained VAE with latent optimization to perform retrieval and assembly jointly, yet relies on hyperparameters or complex rules to control the number of retrieved parts, and does not explicitly model inter-part or part-whole relationships. GPAT leverages a segmented target point cloud to facilitate the assembly process. While it demonstrates the capability to successfully assemble general unseen parts, it still relies on a predefined set of input parts and does not consider the retrieval stage. In contrast, our approach jointly reasons about segmentation, retrieval, and assembly with explicit part-part and part-whole consistency, while supporting variable-length parts output.

\subsection{3D Part Retrieval and Fitting}
The 3D part retrieval and fitting problem aims to reconstruct the target shape by retrieving compatible components from a given library and fitting them to the target. Notably, significant research has focused on 3D primitive fitting and shape abstraction, which utilize basic geometric primitives (e.g., cuboids, cylinders) to represent complex shapes. Researchers have explored various fitting approaches, including optimization-based fitting~\cite{schnabel2007efficient, li2011globfit, liu2023marching}, network-based fitting (unsupervised~\cite{tulsiani2017learning, yang2021unsupervised, Paschalidou2019CVPR} and supervised~\cite{zou20173d}), and more recently, Large Language Model (LLM) based fitting~\cite{tian2025llm}. While primitive representations offer advantages such as low storage and fast rendering, retrieving and assembling components from real-world part libraries more closely mirrors practical manufacturing pipelines and can produce more complex, higher-quality assemblies. However, general part retrieval and assembly remains largely underexplored~\cite{uy2021joint, xu2023unsupervised}, and these approaches do not inherently support variable-length retrieval and depend on hyperparameters or post-processing. LLM-Primitives~\cite{tian2025llm} addresses this via LLM-based approach, but has only been validated on simple primitives, falling short of handling realistic part complexity. Our method focuses on reconstructing complete target shapes by retrieving and assembling parts from a given library and can produce variable-length parts output based on given target.

\subsection{Transformers and Set Prediction}
The Transformer architecture~\cite{vaswani2017attention}, originally designed for sequence modeling, has been widely used in various 2D and 3D tasks, e.g., Vision Transformer~\cite{dosovitskiy2020image} and Point Transformer~\cite{zhao2021point, wu2022point, wu2024point}. Set prediction requires a model to output a variable-size, unordered set of elements, which is inherently difficult for canonical deep learning networks. A representative milestone is DETR~\cite{carion2020end}, which pioneered a query-based transformer framework for variable-length set prediction without hand-crafted heuristics such as anchors or non-maximum suppression (NMS). Building on DETR, this query-based paradigm has been extended to various 3D tasks, such as 3D object detection~\cite{misra2021end, liu2021group} and instance segmentation~\cite{schult2022mask3d, sun2023superpoint}. Inspired by DETR, we adopt a query-based transformer framework to jointly address part retrieval and assembly.

\section{Method}
\label{sec:method}
Given a part library containing canonical-pose candidate part point clouds and a target point cloud (e.g., obtained from image-to-3D models or real scans), our goal is to retrieve a subset of parts from the part library (with repetition allowed) and predict a rigid transformation for each selected part, such that the resulting assembly reconstructs the target shape. In the following, we first describe our network architecture in \autoref{subsec:structure}, followed by the training losses in \autoref{subsec:loss}. Finally, we introduce a segmentation-enhanced post-processing optimization module in \autoref{subsec:post-process}.

\begin{figure*}[t]
    \centering
    \includegraphics[width=\textwidth]{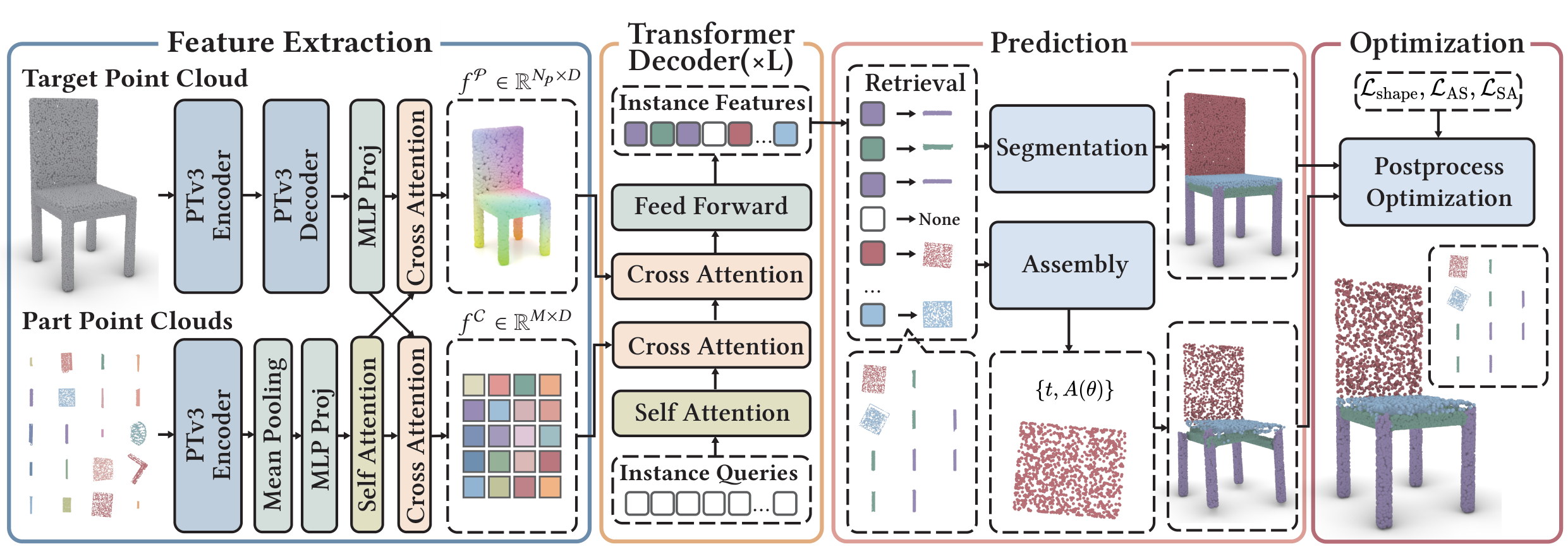}
    \caption{\textbf{Overview of our framework.} Given a target point cloud and a set of candidate parts, we employ Point Transformer v3 layers followed by self- and cross-attention mechanisms to extract point-wise features $f^\mathcal{P}$ and part global features $f^\mathcal{C}$. These representations serve as conditions for the Transformer decoder layers, where a set of learnable instance queries interacts with the encoded features. The decoded instance features are then fed into parallel prediction heads to determine part assignments (classifying into candidate indices or empty class), estimate 6-DoF assembly poses, and generate segmentation masks. Finally, a segmentation-enhanced optimization module refines the assembly configuration to yield the precise result.}
    \Description{pipeline}
    \label{fig:pipeline}
\end{figure*}

\subsection{Network Architecture}
\label{subsec:structure}
Our PART network consists of three stages: the feature extraction, the transformer decoder, and the prediction; see~\autoref{fig:pipeline}.

\paragraph{Features Extraction} For both input part point clouds $\mathcal{C} = \{\mathcal{C}_i\}_{i=1}^M$ and target point cloud $\mathcal{P}$, we use point transformer v3~\cite{wu2024point}, a state-of-the-art backbone network for point cloud deep learning, to extract global features $\in \mathbb{R}^{D}$ of each part and point-wise features $\in \mathbb{R}^{N_p \times D}$ of the target. Here $N_p=10000$ represents the total number of target points, and $D$ denotes the hidden dimension. 
Detailed PTv3 network hyperparameter configuration can be found in the supplementary material. The following self- and cross-attention blocks then exchange information between parts and target, yielding the contextualized part features $f^{\mathcal{C}} = \{f_i\} \in \mathbb{R}^{M\times D}$ and target shape feature $f^{\mathcal{P}}\in \mathbb{R}^{N_p \times D}$.

\paragraph{Transformer Decoder} The number of parts in different assemblies within the same category may vary. For example, chairs can be decomposed into four parts in one design and seven parts in another. This makes the retrieval and assembly task a variable-length set prediction problem. To overcome this challenge, inspired by DETR~\cite{carion2020end}, we employ a query-based transformer decoder architecture. Specifically, we utilize a fixed number of $N$ learnable instance queries $Q^{(0)} = \{Q_j\} \in \mathbb{R}^{N\times D}$ to act as agents to retrieve appropriate parts. Each query is trained to either match with a candidate part or fall into a dedicated \emph{empty} class. This formulation naturally supports variable-length outputs, as queries are allowed to select the \emph{empty} class when no corresponding part exists. Moreover, parts and instance queries are processed in parallel, enabling efficient inference for the input part library. 
The queries are refined by $L$ transformer decoder layers. Within each layer, a query first attends to the other queries (modeling inter-dependencies among different instances) and then sequentially cross-attends to the part features $f^{\mathcal{C}}$ and to the target features $f^\mathcal{P}$:
\begin{equation}
    Q^{(\ell)} = \mathrm{CrossAttn}(\mathrm{CrossAttn}(\mathrm{SelfAttn}(Q^{(\ell-1)}), f^{\mathcal{C}}), f^\mathcal{P}).
\end{equation}
We write the final per-instance features as $f^{Q} \:= \{Q_j^{(L)}\}\in \mathbb{R}^{N\times D}$, which jointly encode the geometry of candidate parts and the structural context of the target.

\paragraph{Prediction} The per-instance features $f^{Q}$ are decoded into three coupled outputs in our prediction head, including part retrieval, target point-cloud segmentation, and assembly pose estimation. 
First, the \textbf{retrieval head} $R$ classifies each query instance $f_j^{Q}$ as one of the $M$ candidate parts plus an \emph{empty} class. 
\begin{equation}
     R = \text{MLP}_{R_Q}(f^Q)\,\text{MLP}_{R_\mathcal{C}}(f^{\mathcal{C}*})^{\!\top}/\sqrt{D} \in\mathbb{R}^{N\times(M+1)},
\end{equation}
where the extended part features are $f^{\mathcal{C}*} = f^{\mathcal{C}} \cup\{e\}$ and $e\in\mathbb{R}^{D}$ is a learnable token for representing the \emph{empty} class. Second, the \textbf{segmentation head} $S$ classifies the target point $f^{\mathcal{P}}$ as one of the $N$ query instances:
\begin{equation}
    S = \text{MLP}_{S_\mathcal{P}}(f^{\mathcal{P}})\,\text{MLP}_{S_Q}(f^Q)^{\!\top}/\sqrt{D} \in\mathbb{R}^{N_p\times N}.
\end{equation}
Lastly, the \textbf{assembly head} predicts a translation $t_j$ and a rotation quaternion $q_j$ per query. 
The pose head utilizes the instance feature $f_j^{Q}$ and a soft max summary of the extended part features:
\begin{equation}
\begin{aligned}
    \big[t_j, A_j\big] &=\text{MLP}_{\text{pose}}\left(\,f^{Q}_{j}\;\Vert\;\mathrm{softmax}(R_j)\,f^{\mathcal{C}*}\right) \;\in\;\mathbb{R}^{3+10}, \\
    q_j &= \argmin_{\|q\|=1,\, q\in \mathbb{R}^{4}} q^\top A_j\, q,
\end{aligned}
\end{equation}
where we follow the A-matrix parameterization~\cite{peretroukhin_so3_2020, lin2023algebraically} for predicting the rotation quaternion to avoid suboptimality, a common issue that arises in $\mathrm{SO}(3)$ rotation regression~\cite{zhou2019continuity, geist2023rotations, xiang2020revisiting}. Specifically, the network first predicts a symmetric matrix $A_j$ with 10 parameters. Then, the predicted quaternion $q_j$ is recovered as the eigenvector of $A_j$ associated with the smallest eigenvalue.

\subsection{Training and Losses}
\label{subsec:loss}
This subsection introduces our training procedure and loss terms. During training, we are given the ground-truth set of parts $\mathcal{C}^{g}$ along with their translations $t^g_k$ and rotation quaternions $q^g_k$, where $k = 1, \cdots, N^g$. To enable a stable training, we perform a bipartite matching to assign each instance query to the most suitable ground-truth part, thereby eliminating the permutation ambiguity inherent in set prediction. Note that we have more queries than the number of ground-truth parts $(N^g \leq N)$. Queries that are not assigned to any ground-truth part are matched to an \emph{empty} class. Our bipartite matching algorithm minimizes the sum of the matching cost $c_{ij}$ between the query $Q_i$ and the ground-truth part $\mathcal{C}^g_j$.
\begin{equation}
    c_{ij} = -\lambda_{1} c^R_{i, j} + \lambda_2 \Vert t_i - t^g_j \Vert_2 - \lambda_3 c^S_{i, j}.
\end{equation}
The first term $c^R_{i, j}$, measures the retrieval match cost. As each ground truth part $\mathcal{C}^g_j$ must have a corresponding part $\mathcal{C}_{j'}$ in the input part library $\mathcal{C}$, this cost term encourages maximizing the retrieval probability $c^R_{i, j} = R(Q_i, \mathcal{C}_{j'})$ of assigning the part $\mathcal{C}_{j'}$ to the query $Q_i$.
The second term penalizes the distance between the predicted and ground-truth part translations. 
The third term $c^S_{i, j}$, measures the segmentation match cost.
For the $j$-th ground-truth part $\mathcal{C}^g_j$, it must correspond to a set of points $\mathcal{P}_j$ in the target $\mathcal{P}$. We define the segmentation cost between query $Q_i$ and ground-truth part $\mathcal{C}^g_j$ as:
\begin{equation}
    c^S_{i,j} = \frac{1}{|\mathcal{P}_j|}\sum_{p \in \mathcal{P}_j} S(p, Q_i) - \frac{1}{|\mathcal{P}\setminus\mathcal{P}_j|}\sum_{p \notin \mathcal{P}_j} S(p, Q_i),
\end{equation}
where two terms respectively encourage maximizing the segmentation probability of $Q_i$ on points inside $\mathcal{P}_j$ and minimizing it on points outside. The hyperparameters are set to $\lambda_1=3, \lambda_2=\lambda_3=1$. More details are provided in the supplementary material. For the $k$-th ground-truth part $\mathcal{C}^g_{k}$, we denote its best match query as $Q_{[k]}$ in the followings.

\begin{figure}
    \centering
    \includegraphics[width=\linewidth]{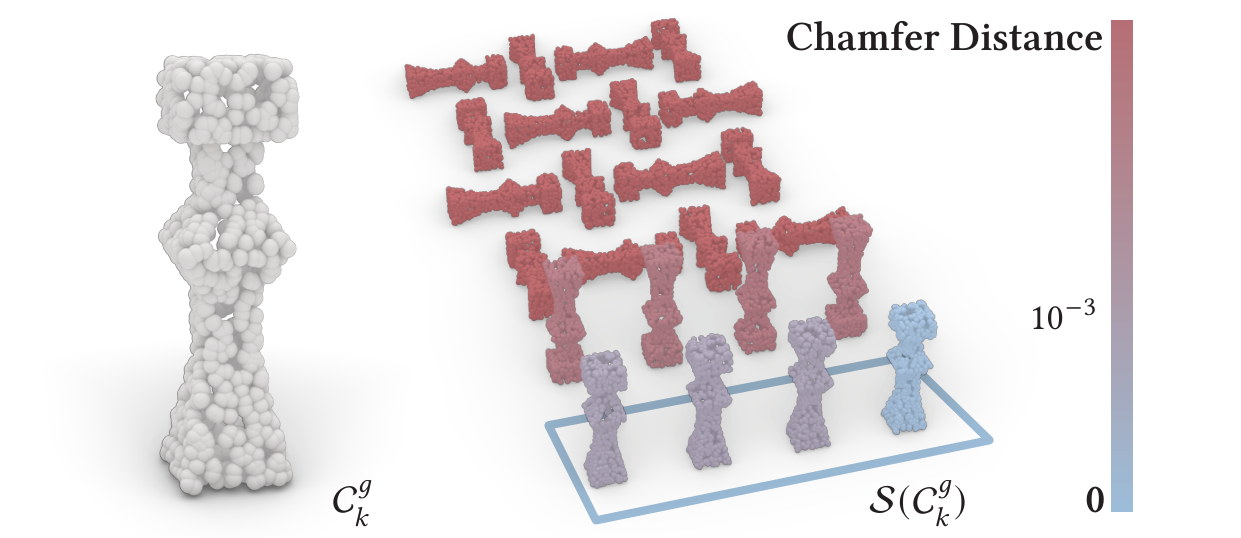}
    \caption{Illustration of part symmetry handling. For a given part $\mathcal{C}^g_k$, we evaluate all 24 discrete rotations. Valid transformations $\mathcal{S}(\mathcal{C}^g_k)$ are selected based on a low Chamfer Distance (colored blue) to the original shape.}
    \Description{sym}
    \label{fig:sym}
\end{figure}

The overall loss contains three tasks: part retrieval, 6-DoF poses for assembly, and point-wise segmentation. The \textbf{pose losses} include translation and rotation loss. Translation is supervised with a standard $\ell_2$ loss:$
    \mathcal{L}_T= \frac{1}{N^g} \sum \| t_{[k]} - t^g_k \|_2^2.$
For rotation loss, we notice that parts might exhibit discrete symmetries (e.g., a cube looks identical under 90° rotations). To account for such symmetries, we construct a valid set of rotation quaternions $\mathcal{S}(\mathcal{C}^g_k)$ by applying the 24 cubic symmetry rotations and retaining only those that yield a small Chamfer distance (i.e., $10^{-3}$) with respect to the ground-truth part, as illustrated in \autoref{fig:sym}. The training loss for part rotation is:

\begin{equation}
\begin{aligned}
    \mathcal{L}_Q &= \frac{1}{N^g}\sum_k\min_{q^s_k\in \mathcal{S}(\mathcal{C}^g_k)} \left( \lambda_{\text{CD}}\mathcal{L}_{\text{CD}}+\lambda_{\text{Geo}}\mathcal{L}_{\text{Geo}} \right), \\
    \mathcal{L}_\text{CD} &= \text{CD}(q_{[k]}q^s_k\mathcal{C}^g_k, q_k^g\mathcal{C}^g_k), \\
    \mathcal{L}_\text{Geo} &= \frac{1}{\pi}\arccos (2\langle q_{[k]}q^s_k, \, q^g_k \rangle^2-1).
\end{aligned}
\end{equation}
The loss term $\mathcal{L}_\text{CD}$ measures the Chamfer Distance between the point clouds transformed using the predicted rotation $q_{[k]}q_k^s$ and the ground truth rotation $q_k^g$. The loss term $\mathcal{L}_\text{Geo}$ directly penalizes the geodesic distance of the two rotations. The two terms are complementary: $\mathcal{L}_\text{CD}$ offers a smooth, spatially grounded gradient, while $\mathcal{L}_\text{Geo}$ prevents convergence to orientations that are metrically close in point-cloud space but far in rotation space.

In addition, in order to provide supervision for the overall shape of each part and the part retrieval scores, we introduce the following soft chamfer loss:
\begin{equation}
    \mathcal{L}_C = \frac{1}{N^g}\sum_{k=1}^{N^g} \sum_{j=1}^{M+1} \hat{S}_{[k], j} \text{CD}(q_{[k]}\mathcal{C}_j+t_{[k]}, \mathcal{C}^g_{k}),
\end{equation}
where $\hat S_{[k], j}$ is the softmax retrieval probability over $M$ candidate parts $\{\mathcal{C}_j\}_{j=1}^M$ plus an \emph{empty} class. This loss back-propagates shape reconstruction error through the retrieval distribution, encouraging the network to assign high probability to the candidate whose transformed geometry best matches the ground truth.

Since both retrieval and segmentation can be formulated as classification tasks, we employ cross-entropy loss to directly supervise the loss between the predicted scores $R,S$ and the corresponding matched ground truth labels $R^g, S^g$: $\mathcal{L}_{R} = \text{CE}\left(R,R^g \right), \mathcal{L}_{S} = \text{CE}\left(S,S^g \right)$. Finally, the overall loss function is computed as:
\begin{equation}
	\mathcal{L} = \lambda_{T}\mathcal{L}_T + \mathcal{L}_{Q} + \lambda_{C}\mathcal{L}_C + \lambda_{R}\mathcal{L}_R + \lambda_{S}\mathcal{L}_S,
\end{equation}
here we set hyperparameters as $\lambda_{T}=1, \lambda_{\text{CD}}=10, \lambda_{\text{Geo}}=0.1, \lambda_{C}=10, \lambda_{R}=1, \lambda_{S}=0.1$.

\subsection{Inference and Post-process Optimization}
\label{subsec:post-process}
This subsection describes our inference procedure and the subsequent post-processing optimization. During inference, we omit the bipartite matching step and directly assign each instance query to either a candidate part or the \emph{empty} class by selecting the maximum-probability index from the retrieval head output. Each selected part is additionally associated with a predicted 6-DoF pose and point-wise segmentation logits for the following post-process optimization.

Suppose the network outputs segmentation logits $S$, along with a set of $M_p$ selected parts ${\hat{\mathcal{C}}_k}$ and their predicted poses ${(\hat{t}_k, \hat{q}_k)}$. We denote the transformed point cloud of each selected part as $\hat{\mathcal{P}}_k = \hat{q}_k \hat{\mathcal{C}}_k + \hat{t}_k$. Our key observation is an \emph{assembly--segmentation duality}. On the one hand, transforming the selected parts using the predicted poses induces a refined segmentation of the target point cloud, denoted by $\hat{S}$. On the other hand, the predicted point-wise segmentation logits define a set of segmented target point clouds $\{\mathcal{P}^S_k\}$, which provides strong geometric cues about the position and orientation of the corresponding selected part $\hat{\mathcal{C}}_k$. We therefore refine poses and segmentation jointly, driving them towards mutual consistency while keeping the assembled shape faithful to the target:

\begin{equation}
\begin{aligned}    
    & \min_{\hat{t}_k, \hat{q}_k, S}\;
    \lambda_{\text{Shape}}\mathcal{L}_{\text{Shape}}
    +\lambda_{SA}\mathcal{L}_{SA}
    +\lambda_{AS}\mathcal{L}_{AS}, \\
    &\mathcal{L}_{\text{Shape}} =\text{CD}\left({\textstyle \bigcup} \hat{\mathcal{P}_k},\mathcal{P}\right), \\
    &\mathcal{L}_{SA} =\frac{1}{M_p}\sum_k \text{CD}(\hat{\mathcal{P}}_k, \mathcal{P}^S_k), \quad 
    \mathcal{L}_{AS} =\text{CE}(S, \hat{S}),
\end{aligned}
\end{equation}
where we set $\lambda_{\text{Shape}}=\lambda_{\text{SA}}=10$, $\lambda_{\text{AS}}=0.1$. The first objective term, $\mathcal{L}_{\text{Shape}}$, minimizes the Chamfer distance between the union of transformed parts and the target shape. The second term, $\mathcal{L}_{SA}$, leverages the predicted segmentation to refine pose estimation, where $\mathcal{P}^S_k$ denotes the segmented target point cloud associated with part $k$ according to the predicted segmentation logits $S$. The final term, $\mathcal{L}_{AS}$, uses the predicted part poses to refine the segmentation. $\hat{S}$ is computed as the segmentation induced by assigning each target point to the transformed part with the nearest point correspondence. 

To mitigate local minima on the non-convex rotation manifold, we run the optimization in parallel from all $24$ symmetry-equivalent rotations of $\hat{q}_k$ and keep the one with the lowest final loss. Each branch runs for $250$ steps with AdamW~\cite{loshchilov2017decoupled}, a learning rate of $0.1$, and cosine annealing.
\section{Experiments}
\label{sec:experiments}

\begin{table*}[t]
\centering
\caption{Quantitative comparisons with assembly methods. We report the assembly conditions for each method, highlighting the \textbf{best} and \underline{second best} results.}
\label{tab:comparison_partnet}
\small
\begin{tabular}{lc|cccc|cccc|cccc}
\toprule
  & & \multicolumn{4}{c|}{Chair} & \multicolumn{4}{c|}{Table} & \multicolumn{4}{c}{Lamp} \\
\midrule
Method & Condition & SCD $\downarrow$ & PA $\uparrow$ & CA $\uparrow$ & SR $\uparrow$ & SCD $\downarrow$ & PA $\uparrow$ & CA $\uparrow$ & SR $\uparrow$ & SCD $\downarrow$ & PA $\uparrow$ & CA $\uparrow$ & SR $\uparrow$ \\
\midrule
DGL~\cite{zhan2020generative} & - &  9.62 & 37.50 & 22.56 & 8.06 & 5.23 & 47.44 & 36.65 & 20.13 & 7.84 & 32.70 & 38.23 & 14.18 \\
RGL~\cite{narayan2022rgl} & Sequence &  10.10 & 44.25 & 28.90 & 7.98 & 5.29 & 49.80 & 39.45 & 20.97 & 8.71 & 31.30 & 48.58 & 14.90 \\
GPAT~\cite{li2023rearrangement} & Point Cloud & 7.78 & 58.41 & 35.39 & 20.38 & 7.73 & 51.79 & 26.55 & 20.37 & 8.23 & \underline{60.58} & 34.41 & 26.12 \\
ScorePA~\cite{cheng2023score} & - & 7.36 & 41.76 & 28.25 & 8.72 & 4.53 & 51.52 & 39.82 & 17.71 & 8.66 & 31.21 & 48.45 & 12.29 \\
3DHPA~\cite{du2024generative} & - & \underline{5.52} & 62.34 & 47.18 & 20.24 & 4.85 & 54.36 & 48.95 & 28.75 & \underline{7.65} & 36.35 & 57.28 & 18.92 \\
SPAFormer~\cite{xu2025spaformer} & Sequence & 6.23 & 55.83 & 38.73 & 24.57 & \underline{3.95} & 64.43 & \underline{61.01} & 39.69 & 11.34 & 36.92 & 47.19 & 18.14 \\
Assembler~\cite{zhao2025assembler} & Image & 7.21 & \underline{66.17} & \underline{52.20} & \underline{24.92} & 4.72 & \underline{70.44} & 57.72 & \underline{43.46} & 16.24 & 51.42 & \underline{57.44} & \underline{33.24} \\
\midrule
Ours-PN & Point Cloud & 6.63 & 59.26 & 43.05 & 14.25 & 4.44 & 61.61 & 58.26 & 29.55 & 10.24 & 42.57 & 55.64 & 16.62 \\
Ours & Point Cloud & \textbf{1.19} & \textbf{86.44} & \textbf{69.90} & \textbf{49.86} & \textbf{0.72} & \textbf{84.71} & \textbf{75.95} & \textbf{59.96} & \textbf{1.63} & \textbf{75.74} & \textbf{58.37} & \textbf{52.24} \\
\bottomrule
\end{tabular}
\end{table*}

\begin{figure*}
    \centering
    \includegraphics[width=0.98\textwidth]{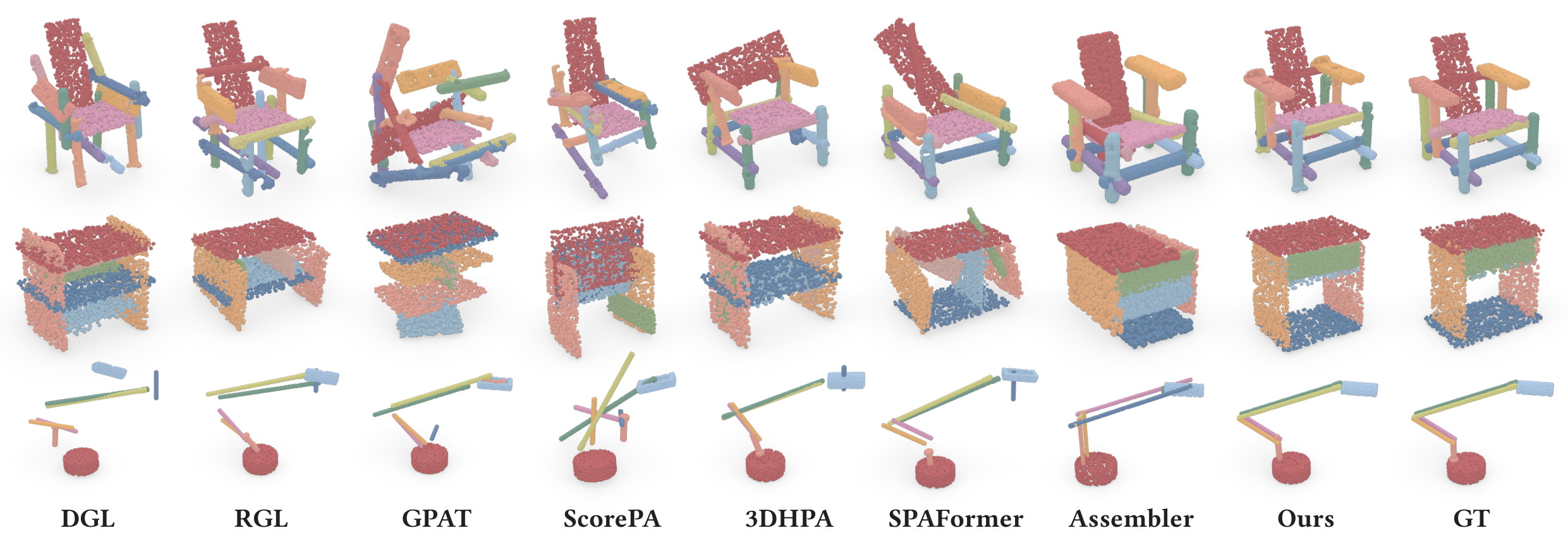}
    \caption{Qualitative Comparisons with assembly methods.}
    \label{fig:comparison}
\end{figure*}

\subsection{Datasets}
\paragraph{PartNet} Following previous works~\cite{zhan2020generative, xu2025spaformer}, we benchmark our method on PartNet~\cite{mo2019partnet} across the \textit{Chair}, \textit{Table}, and \textit{Lamp} categories at the finest-grained segmentation level, filtering for objects with 2 to 20 parts. We adopt the official train/validation/test splits of 70\%/10\%/20\% and report results under the same evaluation protocol for fair comparison.

\paragraph{Curated Data} To extend our framework beyond these canonical categories to general object retrieval and assembly, we collect and process 80K+ data from multiple part-annotated 3D datasets, including PartNet~\cite{mo2019partnet}, PartNeXt~\cite{wang2025partnext}, 3DCoMPaT\texttt{++}~\cite{slim20253dcompat++} and PartVerse-XL~\cite{ding2025fullpart}.

\paragraph{3D-FRONT} We also validate our model on the 3D-FRONT~\cite{fu20213d} dataset to evaluate its performance on scene-level assembly, where the goal is to arrange multiple furniture objects into a coherent room layout.

\subsection{Implementation Details}
\label{subsec:details}
\paragraph{Configurations} We provide two model configurations tailored to different task scales. For category-specific benchmarks on PartNet and scene-level assembly on 3D-FRONT, we set the hidden dimension $D$ to 384, yielding a model of 51.9M parameters. For general object assembly on the curated data, we set $D$ to 768, resulting in 113M parameters. All models are trained on 8 NVIDIA A800 GPUs using AdamW~\cite{loshchilov2017decoupled} with a batch size 16 per GPU, learning rate of $8 \times 10^{-5}$, weight decay of $10^{-4}$, and cosine annealing schedule. Further details on datasets and training are provided in the supplementary material.

\paragraph{Train Short, Test Long} A key practical question is how to choose the training-time library size $M$. A large $M$ inflates the quadratic attention cost, slows each iteration, and makes convergence noticeably harder. We therefore adopt a \emph{train-short-test-long} strategy: during training each shape is paired with a fixed-size library of $M{=}20$ parts, comprising its own ground-truth parts (deduplicated by geometric identity) and distractor parts randomly sampled from other shapes in the training set; at inference time the same model is applied to libraries of arbitrary size $M'\gg 20$ (by default we set $M'=100$ in the experiments). This strategy is effective for two reasons. First, candidate parts are encoded as an unordered set without positional encodings, making the number of attention keys a runtime variable that the architecture is inherently agnostic to~\cite{press2021train, kool2018attention, kazemnejad2023impact}. Second, resampling distractors at every iteration exposes the model to thousands of distinct library compositions, covering a negative pool far larger than $M$. We empirically verify this strategy in \autoref{subsec:more}.

\subsection{Assembly Comparisons on PartNet Dataset}
\paragraph{Comparisons with Assembly Methods} We first evaluate our method against existing approaches that assemble a fixed set of given parts. Following previous works, we conduct the experiments on \textit{Chair}, \textit{Table}, \textit{Lamp} categories from the PartNet~\cite{mo2019partnet} with fixed parts as input, using Shape Chamfer Distance (SCD), Part Accuracy (PA), Connectivity Accuracy (CA), and Success Rate (SR) as evaluation metrics (details are provided in the supplementary material). Since previous approaches mostly use PointNet/PointNet++ as their backbone and do not employ post-processing optimization, we additionally provide a variant (Ours-PN) using PointNet++ without optimization for fair comparison. \autoref{fig:comparison} visualizes the qualitative results, and \autoref{tab:comparison_partnet} summarizes the quantitative performance. The results demonstrate that our 3D target point cloud prior based approach achieves superior assembly accuracy compared to other fixed-part assembly methods. Moreover, Ours-PN outperforms earlier methods (such as DGL and RGL), though it does not show a significant advantage over recent methods like 3DHPA and SPAFormer. We conjecture that these methods benefit from exploiting the fixed, known part set through super-part relation reasoning or order encoding, a prior unavailable in our library-based harder setting. Notably, our full method surpasses all baselines, particularly on the SCD metric, as the optimization enables parts to accurately fit the target shape.

\begin{figure}[h]
    \centering
    \includegraphics[width=\linewidth]{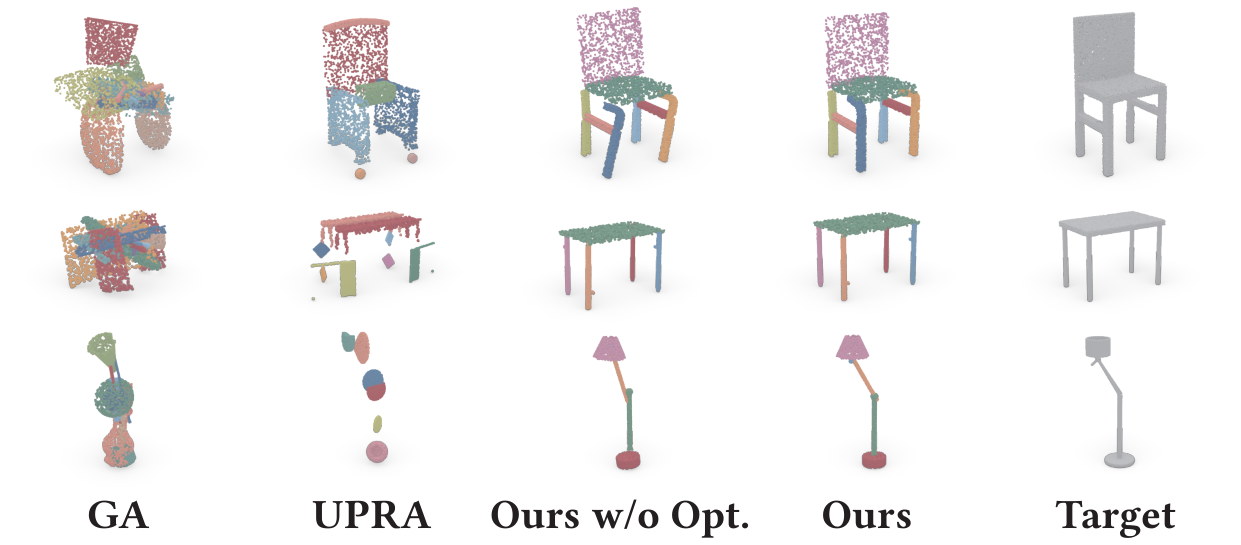}
    \caption{Qualitative comparisons with retrieval and assembly methods. Each shape is reconstructed from a library of 1,000 parts. Our method adaptively handles the variable part count and produces more accurate assembly results combined with segmentation-enhanced optimization.}
    \label{fig:comparison2}
\end{figure}

\paragraph{Comparisons with Retrieval and Assembly Methods} In the second setting, we evaluate our full pipeline on the task of part retrieval and assembly. We compare our approach against two distinct baselines: a Genetic Algorithm (GA) based method, which employs heuristic search and brute-force optimization to align parts onto the target shape, and another baseline UPRA~\cite{xu2023unsupervised}. Since UPRA does not release its training data or pretrained checkpoints, we retrain its model on PartNet using the official codebase. For a fair comparison, all methods use the same candidate library consisting of 1,000 unique parts from the corresponding category's test split (1,000 points per part) and the same 10,000-point target point clouds. As shown in \autoref{tab:comparison_partnet2} and \autoref{fig:comparison2}, our method achieves lower assembly error while requiring significantly less inference time per shape. All methods are benchmarked on a single NVIDIA A800 GPU. Note that the reported inference time does not include data loading.

\begin{table}[t]
\centering
\small
\setlength{\tabcolsep}{4pt}
\caption{Quantitative comparisons with retrieval and assembly methods.}
\label{tab:comparison_partnet2}
\begin{tabular}{l|ccc|ccc}
\toprule
 & \multicolumn{3}{c|}{SCD $\downarrow$} & \multicolumn{3}{c}{Average Time (s)} \\
\midrule
Method & Chair & Table & Lamp & Chair & Table & Lamp \\
\midrule
GA & 8.27 & 6.57 & 7.81 & 99.0 & 71.4 & 101.6  \\
UPRA~\cite{xu2023unsupervised} & 4.46 & 2.29 & 3.24 & 227.6 & 296.0 & 99.1  \\
\midrule
Ours & \textbf{1.84} & \textbf{0.70} & \textbf{3.16} & \textbf{6.43} & \textbf{6.35} & \textbf{4.89} \\
\bottomrule
\end{tabular}
\end{table}

\subsection{Ablation Study}
We conduct ablation studies on the PartNet \textit{Chair} category to validate each component, with results summarized in \autoref{tab:ablation}, where SmIoU (Shape mIoU) serves as the evaluation metric for segmentation quality. We compare our full model against the following baselines: \textit{w/o Optimization} omits the joint optimization stage where assembly and segmentation mutually reinforce each other; \textit{w/o A-Matrix} replaces A-Matrix rotation representation with direct quaternion regression; \textit{w/o Symmetry} removes symmetry matching and uses direct rotation loss; \textit{w/o Fusion} removes the self- and cross-attention based feature fusion between target and part clouds; \textit{Segmentation+PCA} discards all assembly losses and recovers pose indirectly by aligning PCA axes of the predicted segmentation with canonical part poses. Results show that each component contributes positively. In particular, the comparison with \textit{w/o Optimization} confirms that the joint optimization stage improves both assembly and segmentation performance, while \textit{Segmentation+PCA} demonstrates that indirect pose recovery via independent segmentation is inferior to our unified framework, underscoring the necessity of joint training for assembly and segmentation. More ablations are provided in the supplementary material.

\begin{table}[t]
\centering
\caption{Ablation studies on PartNet Chair category. Segmentation metrics are omitted for \textit{w/o A-Matrix} and \textit{w/o Symmetry}, as these are designed for assembly task.}
\label{tab:ablation}
\small
\begin{tabular}{lcccccc}
\toprule
 & SCD $\downarrow$ & PA $\uparrow$ & CA $\uparrow$ & SR $\uparrow$ & SmIoU $\uparrow$ \\
\midrule
w/o Optimization & 3.96 & 80.16 & 61.31 & 33.65 & 74.88   \\
w/o A-Matrix & 4.43 & 78.04 & 56.45 & 31.12 & --  \\
w/o Symmetry & 4.15 & 79.90 & 61.27 & 32.98 & --  \\
w/o Fusion & 4.09 & 80.23 & 60.76 & 33.93 & 75.15  \\
Segmentation+PCA & 5.95 & 63.22 & 32.01 & 23.06 & 64.96 \\
\midrule
Ours & \textbf{1.19} & \textbf{86.44} & \textbf{69.90} & \textbf{49.86} & \textbf{77.16}  \\
\bottomrule
\end{tabular}
\end{table}

\begin{figure*}[t]
    \centering
    \includegraphics[width=\textwidth]{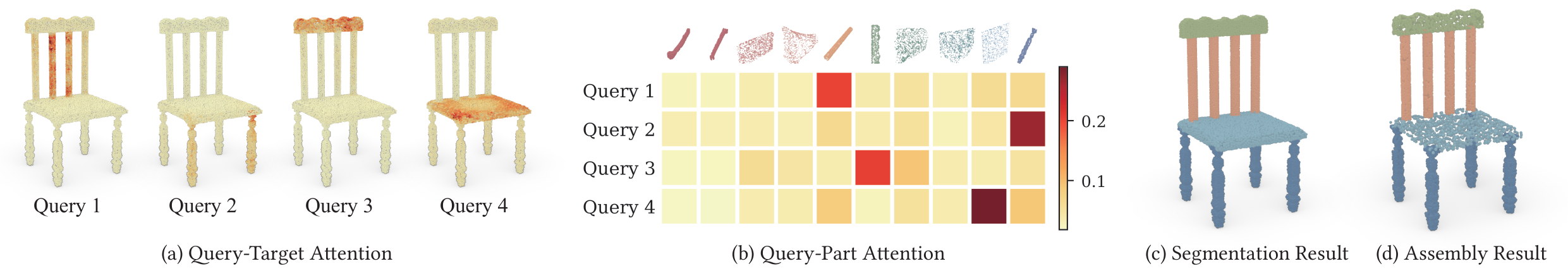}
    \caption{Attention visualization. (a): query-target attention weights. (b): query-parts attention weights. (c)(d): the final segmentation and assembly result.}
    \label{fig:attn}
\end{figure*}

\begin{figure}[h]
    \centering
    \includegraphics[width=\linewidth]{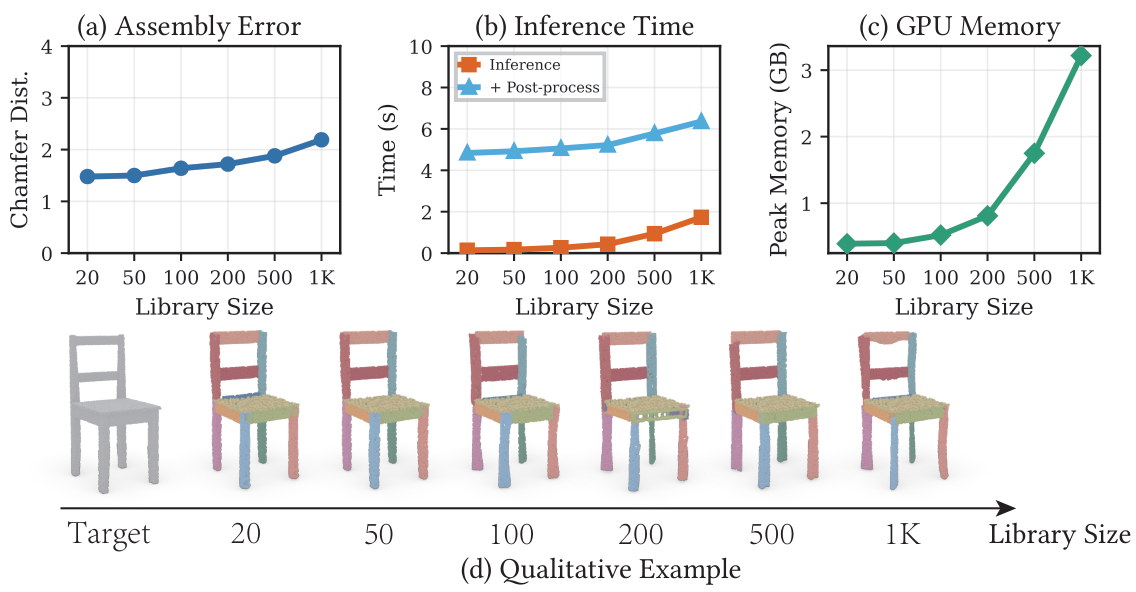}
    \caption{Effect of part library size on assembly performance. (a) Assembly error (SCD). (b) Average inference time. (c) Average GPU memory usage across different library sizes. (d) A qualitative example showing assembly results under varying library sizes.}
    \label{fig:library_scaling}
\end{figure}

\paragraph{Effect of Part Library Size} As discussed before, our query-based transformer architecture naturally accepts a variable number of input parts. To evaluate the effect of library size, we randomly sample 100 shapes from the PartNet \textit{Chair} test set and measure assembly quality, inference time, and GPU memory under library sizes of 20, 50, 100, 200, 500, and 1000 unique parts on a single NVIDIA A800 GPU. As shown in \autoref{fig:library_scaling}, Chamfer distance and inference time increase only marginally as the library grows. Although GPU memory grows due to the quadratic complexity inherent in the transformer's attention mechanism, they remain within a practical range (approximately 3.2GB per shape even at 1,000 parts), which is sufficient for most application scenarios. Qualitatively, despite the increasing retrieval difficulty posed by larger libraries, the assembled shapes maintain strong global consistency with the target, with only minor local deviations. This validates the effectiveness of our train-short-test-long strategy and demonstrates the robustness of the model to varying numbers of input parts.

\subsection{More Analysis}
\label{subsec:more}

\paragraph{Attention Map Visualization} We visualize representative cross-attention weights from the transformer decoder layers in \autoref{fig:attn}. The query-target attention (left) shows that each query attends to a localized region of the target point cloud, indicating that the model learns to decompose the shape into semantically meaningful segments. The query-part attention (middle) reveals that each query identifies the most geometrically compatible part in the library. Through iterative cross-attention, each valid query progressively establishes correspondence between a part and its target placement, yielding consistent segmentation and assembly results (right).

\begin{figure*}
    \centering
    \includegraphics[width=0.98\textwidth]{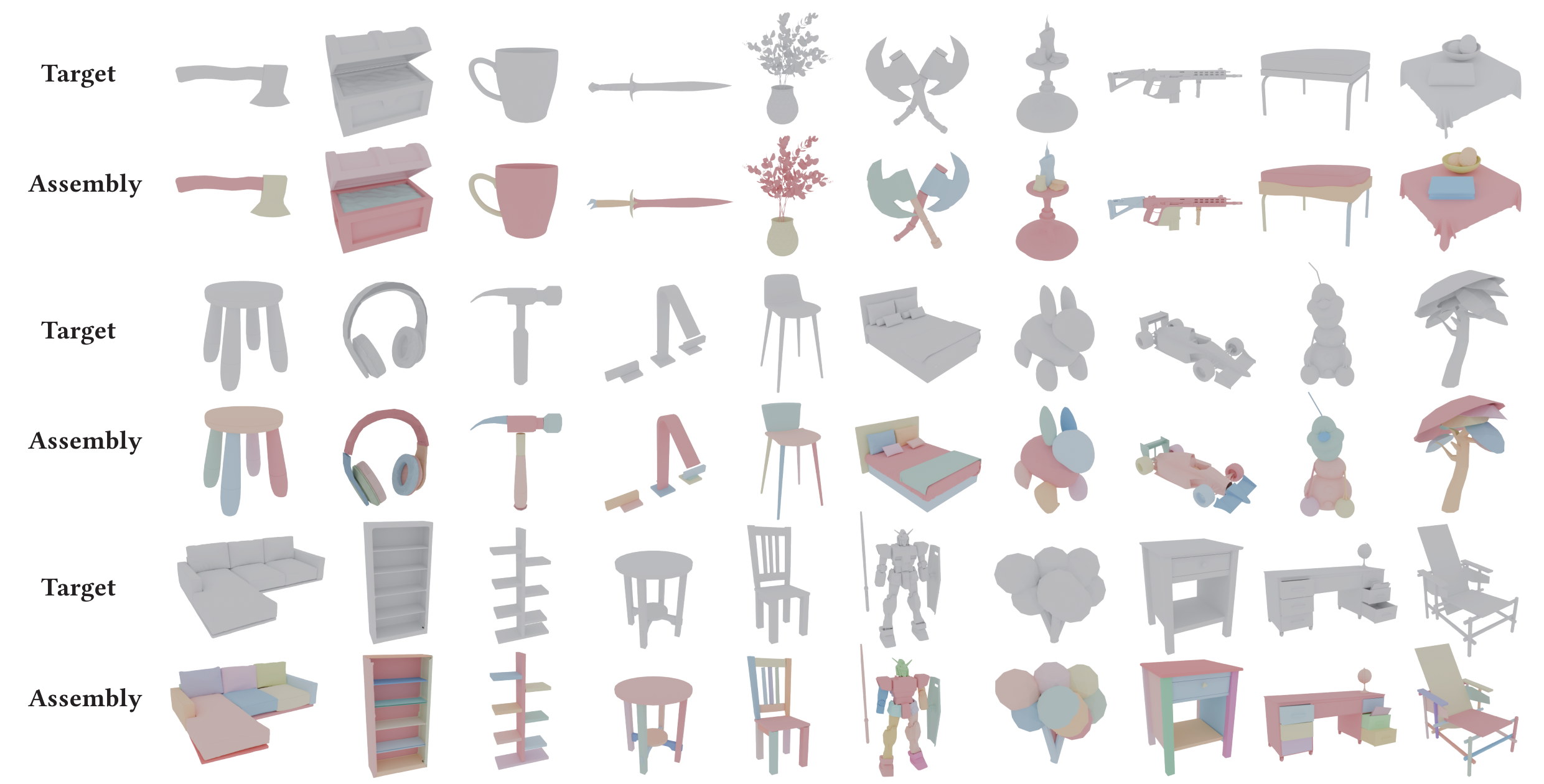}
    \caption{Additional retrieval and assembly results across various categories with varying number of parts. For each shape, the part library contains 100 parts randomly sampled from the test split (including the needed parts). PART is able to retrieve an appropriate number of parts conditioned on the target and assemble them accurately.}
    \label{fig:gallery}
\end{figure*}

\paragraph{Additional Results} \autoref{fig:gallery} visualizes more assembly results on our curated test data. Our framework adaptively retrieves an appropriate number of parts based on the target, producing faithful assemblies ranging from simple to complex geometries. 

\begin{figure}[h]
    \centering
    \includegraphics[width=\linewidth]{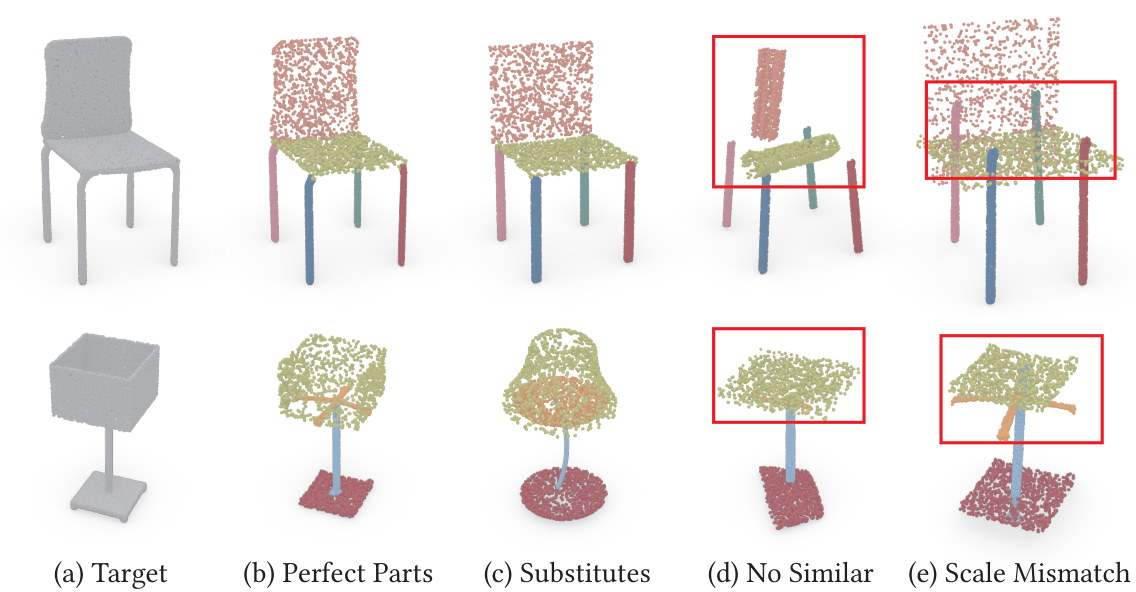}
    \caption{Assembly with/without perfect parts and failure cases. (a): Input target point cloud. (b) Assembly with perfect parts. (c) Assembly without perfect parts, our method still retrieves plausible substitutes. (d) No geometrically similar parts in the library. (e) The parts and target differ in scale. }
    \label{fig:failure_cases}
\end{figure}

\paragraph{Assembly w/o Perfect Parts \& Failure Cases} We further evaluate our model under the challenging setting where the part library does not contain the perfect matched parts (i.e., ground-truth parts of the target shape). As shown in \autoref{fig:failure_cases}, the model can still retrieve geometrically plausible substitutes that closely approximate the target (column 3). However, failure cases arise when the parts in the library differ significantly in scale from the target (column 5) or when there are no geometrically similar parts available (column 4).

\subsection{Exploratory Extensions}

We further demonstrate the scalability of our framework through three exploratory extensions.

\begin{figure*}[t]
    \centering
    \includegraphics[width=\textwidth]{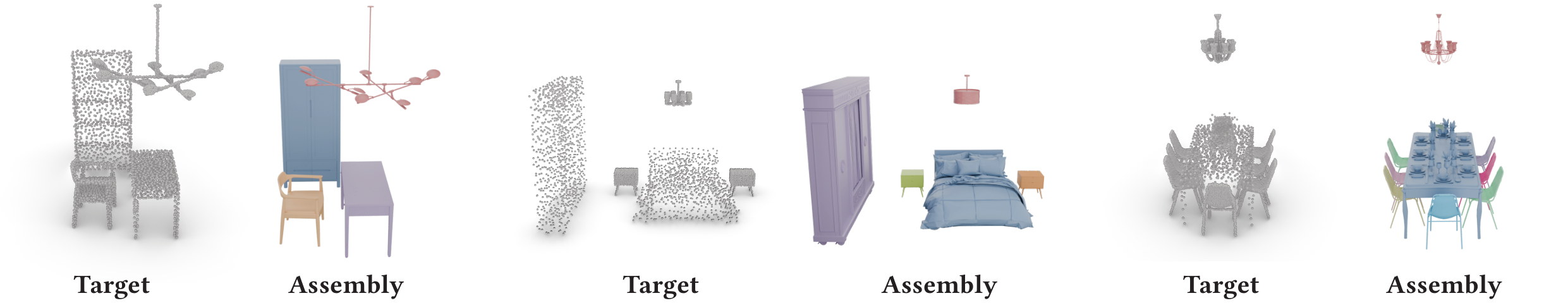}
    \caption{Scene assembly visualizations. Left: Input target point cloud. Right: Our assembled result.}
    \label{fig:scene}
\end{figure*}

\begin{figure*}[t]
    \centering
    \includegraphics[width=0.98\textwidth]{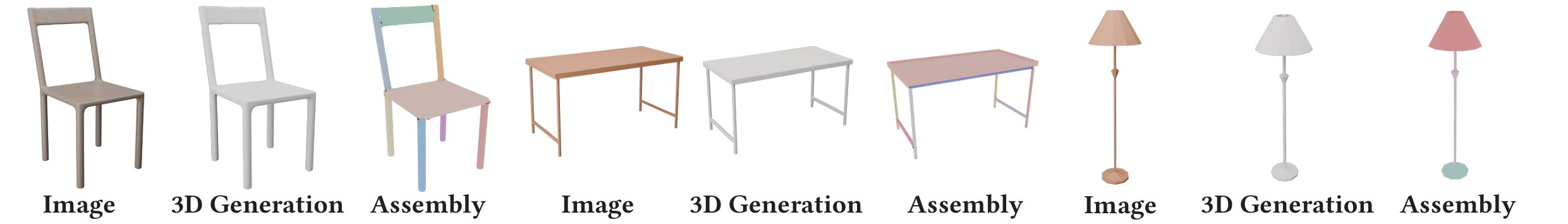}
    \caption{Examples of image-based assemblies. Given a single image as reference (left), we first lift it to a 3D model using Hunyuan3D (middle). This model is then sampled into a point cloud and utilized as the target to obtain the final assembly (right).}
    \label{fig:image_assembly}
\end{figure*}

\paragraph{Image-based Assembly} First, we extend it to the image-based retrieval and assembly by leveraging recent advances in 3D generation (\autoref{fig:image_assembly}). Specifically, we utilize Hunyuan3D~\cite{zhao2025hunyuan3d, lai2025hunyuan3d} to lift single-view images into 3D meshes, from which we sample 10,000 surface points as the input target point cloud. To handle the scale inconsistency between the generated target and the input parts, we normalize the target to a unit sphere and incorporate object scale as an optimizable parameter during post-processing.

\paragraph{Scene-level Assembly} Second, for scene-level assembly on the 3D-FRONT dataset, given an input scene point cloud and a set of candidate assets, our model automatically retrieves and places the appropriate assets to reconstruct the target scene (\autoref{fig:scene}).

\begin{figure}[h]
    \centering
    \includegraphics[width=\linewidth]{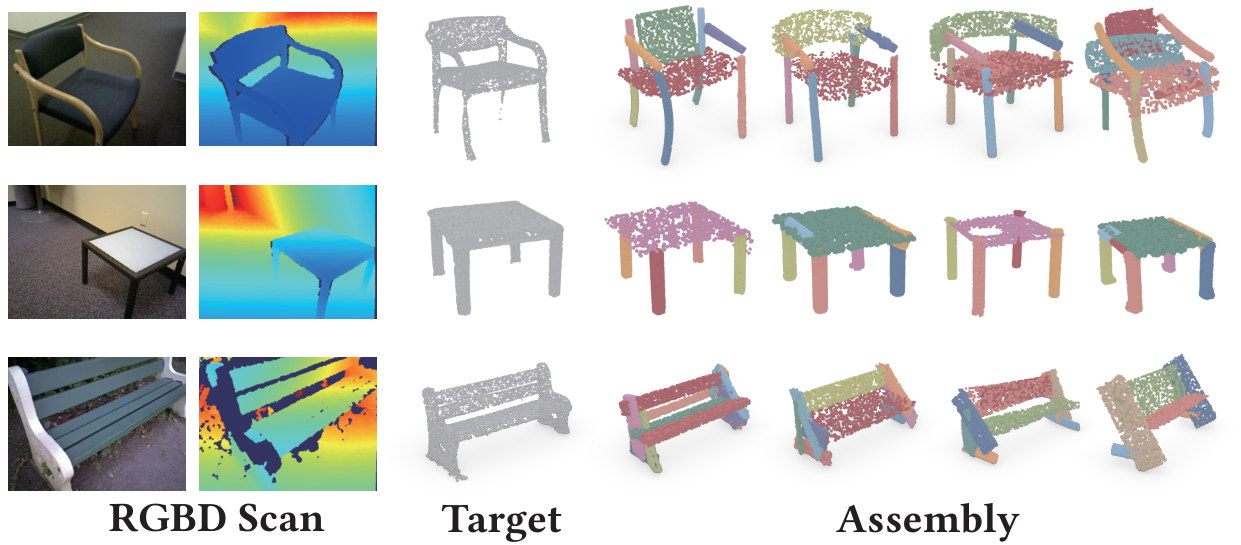}
    \caption{Generalization to real-world scans. Left: Raw RGBD captures from Redwood dataset. Middle: Target point cloud sampled from the scanned model. Right: Diverse assembly results using parts retrieved from the PartNet library to reconstruct the targets.}
    \label{fig:scan}
\end{figure}

\paragraph{Generalization to Real-world Scans} Third, to evaluate the real-world applicability, we conduct a zero-shot generalization experiment using scanned point clouds from Redwood dataset~\cite{Choi2016}. For each scan, we randomly sample 100 parts from PartNet as the library. As shown in \autoref{fig:scan}, our method retrieves appropriate parts and produces plausible assemblies that approximate the scanned inputs, demonstrating the zero-shot generalizability of our framework to real-world inputs. Additional results are provided in the supplementary material.

\section{Conclusion}
We present PART, a unified framework for 3D part retrieval and assembly. By leveraging a transformer-based architecture, our method jointly retrieves parts from a candidate library and estimates their 6-DoF poses in a single forward pass, successfully handling variable-length outputs. Moreover, the joint learning of assembly and segmentation, together with a novel segmentation-enhanced optimization module, further improves the assembly accuracy. Extensive experiments on scene assembly and single-view image-guided assembly further demonstrate the scalability and potential of our framework.

Our work has several limitations. For example, we do not consider inter-part connections and physical constraints such as collision avoidance and structural stability. Additionally, our method assumes rigid transformations and predicts only 6-DoF poses for each part, without accounting for part scaling or non-rigid deformations. For future work, incorporating physics-aware assembly constraints, retrieving and adapting parametric CAD parts, and leveraging generative models or large language models (LLMs) are promising directions to explore.

\begin{acks}
We thank the anonymous reviewers for their constructive comments and valuable suggestions. This work was supported by the National Natural Science Foundation of China (62025207), the BYD-HKUST Joint Lab Research Project (BYD26IS02), and the Shenzhen Loop Area Institute (FPF10120250006).
\end{acks}

\bibliographystyle{ACM-Reference-Format}
\bibliography{reference}

\end{document}